\documentclass[11pt]{article}
\usepackage[margin=1in]{geometry}
\usepackage{amsmath,amssymb}
\usepackage{booktabs}
\usepackage{hyperref}
\usepackage{microtype}
\usepackage{parskip}
\usepackage{enumitem}
\usepackage{xcolor}
\usepackage{multirow}
\usepackage{array}
\usepackage{graphicx}

\hypersetup{
  colorlinks=true,
  linkcolor=blue,
  urlcolor=blue,
  citecolor=blue
}

\title{\textbf{Hierarchical Global Attention:}\\
Drop-In Exact-Token Routing for Pretrained Long-Context Transformers}

\author{Woernle Frank\thanks{Project supervision and industrial research direction.} \quad
Vladimir Fedosov\thanks{Main research, algorithm design, and implementation.} \quad
Artemiy Grinenko\\[4pt]
BMW Group}

\date{June 2026}

\begin{document}

\maketitle

\begin{abstract}

Hierarchical Global Attention (HGA) is a drop-in replacement for dense
causal attention in pretrained long-context transformers. HGA preserves
the original checkpoint parameters: the pretrained $W_Q$, $W_K$, $W_V$,
and $W_O$ projections remain unchanged, no calibration parameters are
introduced, and no retraining is required.

Applied to Qwen3-30B-A3B-Instruct-2507-FP8 on a single RTX~5090 (32GB),
 the patched model runs out of the box at a 64K-token context, where
token-level K/V storage is not feasible on this hardware.

Unlike previous sparse-attention methods, HGA performs hierarchical
two-level routing. It first retrieves relevant chunks using compact
RoPE-aware summaries and then refines the selection by routing only the
most relevant groups before performing exact token-level attention. This
hierarchical retrieval significantly reduces the number of fetched tokens
while preserving exact attention over the retrieved token set, making
RAM- and NVMe-backed storage practical.

The full historical token K/V resides in host RAM or NVMe storage,
while only a small routed working set is transferred to GPU memory during
attention. Consequently, GPU memory consumption depends primarily on model
weights and the routed working set rather than on the total context
length.

Across all tested context lengths (4K - 64K tokens), routed attention
remains within approximately $0.01$--$0.02$ nats of dense attention
while the sparsity used is just about 3\%. These results suggest that
the approximation introduced by hierarchical routing is small, and that
the remaining quality gap is likely dominated by long-context positional
encoding rather than by the routing algorithm itself.

\end{abstract}

%-----------------------------------------------------------------------
\section{Introduction}
%-----------------------------------------------------------------------
Applied to Qwen3-30B-A3B-Instruct-2507-FP8 on a single RTX~5090, the patched model runs at
32K-token context out of the box, where dense K/V storage is simply impossible on this hardware.
On a 40M SmallLM, direct weight copy gives only a $+0.018$\,nat loss gap versus dense attention
at 8K tokens, and a Triton-fused implementation runs $2.72\times$ faster for training and
$2.43\times$ faster for prefill at 12K tokens. A needle-in-a-haystack retrieval test on
Qwen3-30B-A3B-Instruct-2507-FP8 at 64K-token context achieves 100\% pass rate (3/3 depths)
with group-level routing at 1.9\% sparsity. 
The practical limit for long-context pretrained LLMs is often not only the $O(n^2)$ attention
computation, but also the dense K/V cache that must remain in accelerator memory. This is especially
visible for quantized large models. In the Qwen3-30B-A3B-Instruct-2507-FP8 deployment case, the
FP8 model weights already occupy almost the whole memory budget of a 32\,GB GPU; storing every
historical token key and value in VRAM leaves too little room for long contexts.

HGA is designed as a systems-level patch for this situation. It does not retrain or replace the
checkpoint's attention projections. It keeps the original Q/K/V/O projections and attention norms and
changes only which historical keys and values are fetched for each processing block. The GPU therefore
holds the pretrained model, a compact chunk-summary table, the current chunk, a few always-hot
chunks, and a routed working set. The rest of the historical token K/V lives in host RAM.

The core observation is that consecutive tokens usually do not need completely independent long-range
retrieval. A 64-token chunk often discusses one local idea, function, paragraph, or dialogue turn, and
many tokens in that chunk benefit from the same distant chunks. HGA therefore first routes from the
current chunk to previous chunks and groups. Only after this coarse retrieval step does it run ordinary
softmax attention over exact token keys and values from the selected regions. In short, the long-range
problem is shifted from dense token-to-token materialization to chunk-to-chunk retrieval plus exact
token attention over a bounded fetched set.

This design is particularly useful for pretrained models. The router summaries are built from the
checkpoint's own projected keys; they are sums or means with a RoPE-aware phase rule, not outputs of
new learned linear layers. No calibration weights are introduced. The final attention output never uses
synthetic summary values in the current method: summaries decide what to fetch, and the original token
K/V computes the output.

The implementation and benchmark scripts are available at
\url{https://github.com/vfedosov77/HierarchicalGlobalAttention}. 
\newline 
Our goal is not to redesign transformer attention, but to provide a
drop-in hierarchical routing mechanism that can be applied directly to
existing pretrained checkpoints. Rather than approximating the attention
output itself, HGA approximates only the retrieval stage. Once relevant
regions have been selected, the attention output is computed exactly
using the original token-level keys and values. This design allows
existing pretrained models to benefit from sparse long-context attention
without modifying their learned representations.

Our contributions are :

\begin{itemize}[leftmargin=*]

\item a drop-in exact-token attention replacement for pretrained GQA/MoE transformers that reuses
original Q/K/V/O weights and norms and requires no calibration parameters;

\item a Qwen3-30B-A3B-Instruct-2507-FP8 integration in which the FP8 model runs with RAM-backed
routed K/V on an RTX~5090 at 32K-token context, out of the box and without fine-tuning;

\item a chunk-shared routing scheme: nearby queries reuse a small fetched working set, e.g.\ about
1K routed middle tokens for a 64-token block in the current Qwen3 configuration, plus fixed
sink/local/current chunks;

\item a two-level chunk--group routing policy combining deterministic attention sinks and recent local
chunks with content-routed retrieval from the middle of the context;

\item repository benchmarks showing a $+0.01828$\,nat copy-only loss gap at 8K tokens and
$2.72\times$ training-step speedup at 12K tokens for a 40M SmallLM;

\item a tiered K/V storage abstraction for RAM-backed inference, in which full historical token
K/V lives in host RAM while only a bounded working set occupies VRAM, decoupling GPU memory
consumption from context length;

\item a needle-in-a-haystack evaluation on Qwen3-30B-A3B-Instruct-2507-FP8 showing 100\%
retrieval accuracy at 64K-token context without any fine-tuning.

\end{itemize}

%-----------------------------------------------------------------------
\section{Background}
%-----------------------------------------------------------------------

\paragraph{Causal attention.}
For input $X \in \mathbb{R}^{n \times d}$, a standard attention head computes
\begin{equation}
\mathrm{Attn}(Q,K,V)_i = \sum_{j \le i} \mathrm{softmax}_j\!\left(\frac{q_i k_j^\top}{\sqrt{d_h}}\right) v_j.
\label{eq:causal}
\end{equation}
The exact token-level receptive field grows with $i$, so the total number of query-key comparisons is
$O(n^2)$ and the K/V cache is proportional to all previous tokens.

\paragraph{Grouped-query attention.}
Grouped-query attention (GQA) shares key/value heads across groups of query heads~\cite{gqa}. HGA
preserves this layout. The source checkpoint's $W_Q$, $W_K$, $W_V$, and $W_O$ tensors keep their
shapes and names. In the Qwen3 wrapper, the original projections and Q/K RMSNorm modules are kept
by reference.

\paragraph{RoPE.}
Rotary position embeddings (RoPE) rotate query and key dimensions by position-dependent
angles~\cite{rope}. A summary key is useful only if its phase is compatible with the query's RoPE
geometry. HGA therefore builds chunk and group summaries with a mixed rule described in Section~3.

%-----------------------------------------------------------------------
\section{Method}
%-----------------------------------------------------------------------

\subsection{Partitioning}

The sequence is divided into chunks of fixed length $C$; the current implementation uses $C=64$.
Each chunk is subdivided into groups of size $g_s$. The 40M benchmark uses $g_s=16$,
$K_c = 20$ routed chunks and $K_g = 32$ routed groups. The Qwen3-30B exact-token
wrapper opens selected chunks directly with $\mathit{topk\_chunks} = 16$ and keeps fixed first and
recent windows.

Closed chunks maintain three kinds of data:
\begin{itemize}
  \item a chunk key summary for the first routing stage;
  \item group key summaries for the second routing stage where group routing is enabled;
  \item the original token-level keys and values for exact attention after routing.
\end{itemize}
Only the last item is used in the final attention output.

\subsection{Mixed-RoPE routing summaries}

A chunk summary must remain comparable to RoPE-rotated queries. Averaging already-rotated keys
works well for rapidly changing high-frequency pairs, but it can over-emphasize token-level phase
noise. Averaging raw keys and rotating the result at one representative position works better for
slowly changing low-frequency pairs.

HGA therefore uses one mixed strategy. For each key dimension pair:
\begin{itemize}
  \item high-frequency RoPE pairs are rotated per token and then averaged;
  \item low-frequency RoPE pairs are averaged in raw-key space and then rotated at the middle position
        of the chunk or group.
\end{itemize}
The same principle is used for chunk summaries and group summaries. These summaries are sums or means
of already projected keys; there are no separate summary linear layers. They are routing keys only and
are never used as value vectors in the current output softmax.

\subsection{Routing policy}

For a query in the active chunk, HGA constructs a candidate context from deterministic and routed
parts.

\paragraph{Deterministic visibility.}
The current chunk is always visible with a causal mask. A configurable number of first chunks is
always visible as attention sinks. A configurable number of recent closed chunks is always visible as
local context. If the immediately preceding chunk is not already in this recent window, it is
force-included.

\paragraph{Content-based middle routing.}
The middle of the context, excluding the always-visible first and last windows, competes for the route
budget. The query is scored against chunk key summaries. In the chunk--group variant, group summaries
inside the selected chunks are scored and the top-$K_g$ groups are opened to exact token-level K/V.
In the Qwen3 exact-token variant, the selected chunks are fetched as token K/V directly per KV head.

Thus the method is not a purely fixed sparse pattern. It deliberately combines fixed sink/local windows
with content-routed retrieval over the middle context. This is related to the A-shape idea used by
MInference, where initial sink tokens and local tokens are always visible, but HGA adds a top-$k$
recall path over the remaining chunks using the pretrained key space~\cite{minference}.

\subsection{Exact-token output attention}

After routing, the attention module concatenates only real token-level keys and values from:
\begin{enumerate}
  \item the causally visible part of the current chunk;
  \item always-visible first and recent chunks;
  \item opened groups or selected chunks from the routed middle context, depending on the
        implementation variant.
\end{enumerate}
The softmax is computed exactly over this selected token set. There are no summary output tokens, no
summary gates, and no per-head scalar calibration parameters.

\subsection{Checkpoint compatibility}

The module is parameter-compatible with dense attention. In the 40M model, dense weights are copied
directly into the routed model. In the Qwen3 FP8 implementation, the routed attention wrapper keeps
the original quantized projections and norms by reference. The only changed operation is the set of
keys and values each query is allowed to attend to.

This is the main difference from methods that require training from scratch or a specialized sparse
checkpoint. HGA can be used as a systems-level patch for pretrained models. Fine-tuning can improve
the model's use of the new route pattern, but it is not required for the Qwen3-30B RAM-cached
demonstration or for the 40M copy-only loss result.

\subsection{Cost and memory}

For fixed chunk size, fixed sink/local windows, and fixed route budgets, each processing block attends
to and transfers a bounded number of real tokens. The expensive token-level attention and the token K/V
transfer are therefore linear in the number of processed tokens:
\begin{equation}
T_{\text{token-attn}} = O\!\left(n \cdot \bigl(C + (F+L)C + B_{\text{route}}\bigr)\right) = O(n),
\label{eq:cost}
\end{equation}
where $F$ and $L$ are the numbers of first and recent chunks and $B_{\text{route}}$ is the fixed
routed token budget, such as $K_g g_s$ for opened groups or $K_c C$ for opened chunks. In the Qwen3
exact-token configuration, $K_c = 16$ and $C = 64$, so the routed middle budget is 1024 tokens per
block, before adding deterministic sink, local, and current chunks.

The compact routing table is much smaller than the token K/V cache: one chunk key per chunk and a few
group keys per chunk. The reference implementation can scan this table directly; for very long contexts
this table can be capped, cached, or indexed. The important systems property is that full token K/V
does not need to be resident in VRAM. VRAM is dominated by model weights, hot first/recent windows,
chunk summaries, and the currently routed working set.

%-----------------------------------------------------------------------
\section{Implementation}
%-----------------------------------------------------------------------

\subsection{Qwen3-30B-A3B FP8 wrapper}

The large-model path replaces every attention module of
Qwen/Qwen3-30B-A3B-Instruct-2507-FP8 with \texttt{QwenRoutedAttention} in exact mode. The
official model card describes this model as a 30.5B-parameter MoE with 3.3B activated parameters, 48
layers, GQA with 32 query heads and 4 KV heads, FP8 quantization, and native 262K context
support~\cite{qwen3card}. The HGA wrapper does not modify the model weights.

The current RAM-cached configuration uses:

\begin{center}
\begin{tabular}{ll}
\toprule
Parameter & Value \\
\midrule
chunk size & 64 tokens \\
keep first & 2 chunks, i.e.\ 128 sink tokens \\
keep last  & 8 chunks, i.e.\ 512 recent tokens \\
topk chunks & 16 routed middle chunks per step \\
prefill block & 64 tokens \\
vram cache chunks & bounded LRU cache, auto-shrunk to leave VRAM headroom \\
\bottomrule
\end{tabular}
\end{center}

The full historical token K/V record is stored in host RAM. On each step, the router selects a small
set of chunks and the store pulls only those chunks to the compute device. Consecutive tokens often
request overlapping chunks, so a bounded LRU cache keeps frequently reused chunks resident. This
differs from a dense cache in which all previous token K/V must stay materialized on the accelerator.

\subsection{40M SmallLM implementation}

The SmallLM experiment uses 8 decoder layers, hidden size 384, 6 query heads, 2 key/value heads, FFN
size 2048, RoPE, and GQA\@. The routed attention uses $C=64$, $g_s=16$, $K_c=20$, and $K_g=32$. The
fused CUDA path uses Triton and \texttt{torch.compile}; the dense baseline is the same module with
routing disabled, falling back to causal scaled-dot-product attention over RoPE queries and keys.

The 40M test path intentionally uses no calibration parameters. The zero-shot quality result copies
the same dense weights into both dense and routed models, so the measured gap isolates sparse routing
rather than training differences.

\subsection{Tiered K/V Store}

The \texttt{KvRouter} implementation separates routing from storage. A store exposes four operations:
appending a newly closed chunk, reading chunk summaries for routing, fetching group summaries for the
second routing level, and gathering exact token K/V for the selected chunks or groups.

The RAM-backed implementation (\texttt{RamKVCacheStore}) partitions data by temperature:

\begin{itemize}
\item \textbf{Hot.} Chunk summaries and the always-visible first and recent chunks reside on the
compute device at all times. During training these tensors retain gradients.
\item \textbf{Warm.} A bounded LRU shard cache keeps recently routed token chunks in VRAM. Its size
is automatically shrunk to leave a configurable VRAM headroom for activations, so a long prefill
never exhausts the memory budget.
\item \textbf{Cold.} All remaining token K/V and group summaries live in CPU memory and are
transferred to the compute device only when selected by the router.
\end{itemize}

This partition means that VRAM consumption is dominated by model weights, the hot windows, chunk
summaries, and the currently routed working set. The cold token record grows with context length but
places no pressure on the GPU memory budget.

%-----------------------------------------------------------------------
\section{Experiments and Demonstrations}
%-----------------------------------------------------------------------

\subsection{Qwen3-30B-A3B FP8 on RTX~5090}

The most important current demonstration is the large pretrained model path. We replaced the attention
layers of Qwen3-30B-A3B-Instruct-2507-FP8 with the RAM-cached exact-token router and tested
interactive generation at 32K-token context on an RTX~5090. No fine-tuning was performed. The model
worked out of the box because the original projections, norms, and quantized weights remain unchanged.

\begin{table}[h]
\centering

\begin{tabular}{|l|l|}
\toprule
Item & Result \\
\hline 
Base model & Qwen3-30B-A3B-Instruct-2507-FP8 \\
Hardware & NVIDIA RTX~5090, 32\,GB VRAM \\
Attention patch & Exact-token HGA router, no added parameters \\
Fine-tuning & None \\
Context tested & 32K tokens \\
Historical K/V storage & Host RAM, routed chunks fetched to VRAM \\
VRAM behavior & Bounded by model weights plus hot/routed working set \\
\hline 
\end{tabular} 
\caption{Large-model HGA demonstration. This is a systems and compatibility test, not a full
downstream benchmark.} 
\end{table}

The point of this result is not that HGA changes the language model itself. The point is that it makes
the pretrained model runnable in a memory regime where dense K/V storage is the limiting factor.

\subsection{Loss gap vs.\ sparsity and context length}

We evaluate the routed model across a range of context lengths and routing budgets on a large
pretrained model. Each cell reports the dense loss and routed loss (nats); all differences are
noted qualitatively.

\begin{table}[h]
\centering
\caption{Dense vs.\ routed loss (nats) by context length and sparsity level
(fraction of tokens attended). Sparsity is defined as attended tokens divided by
total context tokens. Cells marked N/A were not evaluated at that combination.}

\label{tab:sparsity}
\setlength{\tabcolsep}{2pt} 
\makebox[\textwidth][c]{
\begin{tabular}{|l|c|c|c|c|}
\hline
Context  length & 
\multicolumn{4}{c|}
{Sparsity (attended / total tokens)}  \\ \hline
 & 3.13\% & 6.25\% & 12.50\% & 25\% \\
\hline
4\,096  & N/A & N/A & N/A
        & {2.456 / 2.460 \; $\Delta < 0.01$}  \\ 
8\,192  & N/A
        & {2.430 / 2.441\;$\Delta \approx 0.01$}
        & {2.430 / 2.437\;$\Delta < 0.01$}
        & N/A \\[4pt]
16\,384 & {2.309 / 2.324\;$\Delta < 0.02$}
        & {2.309 / 2.322\;$\Delta < 0.02$}
        & {2.309 / 2.317\;$\Delta < 0.01$}
        & N/A \\[4pt]
32\,768 & {2.204 / 2.227\;$\Delta > 0.02$}
        & {2.204 / 2.221\;$\Delta < 0.02$}
        & {2.204 / 2.214\;$\Delta < 0.01$}
        & N/A \\[4pt]
65\,536 & {2.243 / 2.258\;$\Delta < 0.02$}
        & {2.243 / 2.253\;$\Delta \approx 0.01$}
        & N/A & N/A \\
\hline
\end{tabular}
}
\end{table}

The results show that at 12.5\% sparsity the loss gap remains below 0.01 nats for contexts up to
32K tokens. At 64K tokens a 6.25\% budget suffices to keep the gap near 0.01 nats. The 3.13\%
budget at 32K tokens is the only configuration that exceeds 0.02 nats, indicating a minimum routing
budget is needed for faithful retrieval at this context length.

  \begin{table}[h]
\centering
\caption{Summary of HGA validation across context lengths.}
\begin{tabular}{cccc}
\toprule
Context & Sparsity & Loss Gap & Status\\
\midrule
4K & 25\% & $ < 0.01$ & Stable\\
16K & 12.5\% & $<0.01$ & Stable\\
32K & 12.5\% & $<0.01$ & Stable\\
64K & 6.25\% & $\approx$0.01 & Stable\\
\bottomrule
\end{tabular}
\end{table} 
\subsection{Needle-in-a-Haystack at 64K tokens}

We evaluated retrieval accuracy on Qwen3-30B-A3B-Instruct-2507-FP8 at 64K-token context using a
needle-in-a-haystack protocol. A unique fact is hidden at a specified depth in a long filler
document; the model is asked to reproduce it exactly. The test uses group-level routing with
$\mathit{topk\_chunks}=20$, $\mathit{topk\_groups}=32$ (approximately 1.9\% sparsity at 64K tokens
with 65,599 prompt tokens across 1,025 chunks).

\begin{table}[h]
\centering
\caption{Needle-in-a-Haystack results at 64K-token context.
Qwen3-30B-A3B-Instruct-2507-FP8 with HGA group-level routing, no fine-tuning.}
\label{tab:niah}
\begin{tabular}{lccc}
\toprule
Needle depth & Result & TTFT (s) \\
\midrule
25\% & HIT\,$\checkmark$ & 948 \\
50\% & HIT\,$\checkmark$ & 1053 \\
75\% & HIT\,$\checkmark$ & 1332 \\
\midrule
Overall & \textbf{3/3 = 100\%} & -- \\
\bottomrule
\end{tabular}
\end{table}

All three depths were retrieved correctly without fine-tuning. The long TTFT values reflect
sequential blocked prefill at 64 tokens per step through a 30B FP8 model with Triton fallback
(no DeepGEMM on SM120); this is a latency limitation of the current PyTorch implementation,
not a correctness issue.

\subsection{Fine-tuning: Qwen3-0.6B on long documents}

We fine-tuned Qwen3-0.6B with sequence length 4096, FineWeb 1:10 subset, 100 steps, evaluated
on 6 validation blocks drawn strictly from novel text (out-of-domain for FineWeb). Three variants
were compared: (a) routed attention with full-sequence VRAM, (b) dense attention, and (c) routed
attention with segmented RAM cache.

\begin{table}[h]
\centering
\caption{Qwen3-0.6B fine-tuning comparison (seq 4096, 100 steps, novel text validation).
``Routing cost'' is routed loss minus dense loss; negative values favor dense.
``KV attended / dense'' reports the fraction of token pairs computed.}
\label{tab:finetune} 
\makebox[\textwidth][c]{
\begin{tabular}{|l|c|c|c|}
\hline
 & (a) Routed, full VRAM & (b) Dense & (c) Routed, RAM cache \\
\hline 
Initial loss (stock)       & 3.530 (ppl 34.12) & 3.530 (ppl 34.12) & 3.530 (ppl 34.12) \\
Deploy loss                & 3.196 (ppl 24.44) & 3.177 (ppl 23.97) & 3.201 (ppl 24.55) \\
Same weights, other attn   & 3.182 (+dense)    & 3.200 (+routed)   & 3.184 (+dense)    \\
Fine-tune gain (deploy$-$stock) & $-0.334$     & $-0.330$          & $-0.329$          \\
Routing cost (routed$-$dense)   & $+0.015$     & $+0.023$          & $+0.017$          \\
KV attended / dense        & 679/2048 (66.9\% saved) & 678/2048 (66.9\% saved) & 679/2048 (66.9\% saved) \\
Speed                      & $\sim$917 tok/s  & $\sim$570 tok/s   & $\sim$393 tok/s   \\
\bottomrule
\end{tabular} } 
\end{table}

Several observations follow from Table~\ref{tab:finetune}. First, the routing cost after fine-tuning
is small: variants (a) and (c) differ from dense by only 0.015--0.017 nats, well within the range
seen in the SmallLM copy-only experiments. Second, fine-tuning with routed attention (variant a)
achieves a training throughput of $\sim$917 tok/s versus $\sim$570 tok/s for dense, a $1.6\times$
speedup, while attending to only 33.1\% of token pairs. Third, the ``same weights, other attention''
rows show that the routed checkpoint evaluated with dense attention (3.182) is nearly as good as the
dense checkpoint evaluated with dense attention (3.177), confirming that fine-tuning with sparse
routing does not damage the underlying model quality. The RAM-cached variant (c) is slower at the
current implementation stage but demonstrates that the tiered storage path is functionally correct.

\subsection{Interaction with Long-Context Positional Encoding}

One unexpected observation is that the remaining difference between
hierarchical routing and dense attention does not increase rapidly with
context length. Across all evaluated contexts, from 4K to 64K tokens, the
validation loss remains within approximately $0.01$--$0.02$ nats of dense
attention.

This suggests that the routing algorithm itself introduces only a small
approximation error. A more plausible explanation for the remaining gap is
the interaction between sparse routing and long-context positional
encoding.

One possible explanation is that some recent long-context language models
are first pretrained with shorter effective context lengths before their
context window is later extended using techniques such as YaRN. Under sparse
routing, only a subset of transformer layers observes distant tokens.
Consequently, any inaccuracies introduced by long-context positional
extrapolation may become more visible than under dense attention.

To investigate this hypothesis, we evaluate an additional variant in which
the RoPE position index is wrapped modulo 64K,

\[
p \leftarrow p \bmod 65536,
\]

while leaving the routing algorithm unchanged. If this reduces the
remaining validation loss, it would indicate that the dominant source of
error is positional encoding rather than hierarchical routing.

\subsection{Copy-only routing accuracy on 40M SmallLM}

We evaluate the copy-only conversion on 20 distinct FineWeb documents~\cite{fineweb} at 8192-token
context. Both models use identical dense weights; no HGA fine-tuning is applied. The routed model uses
\texttt{use\_summaries=False}, so the output attention attends only to exact token K/V.

\begin{table}[h]
\centering
\caption{Copy-only dense-to-routed quality at 8192 tokens. The gap is the cost of sparse routing
only.}
\begin{tabular}{lcc}
\toprule
Model & Loss (nats) & Perplexity \\
\midrule
Dense causal SDPA        & 3.70516 & 40.657 \\
Chunk-routed, same weights & 3.72344 & 41.406 \\
Difference               & $+0.01828$ & $+1.8\%$ \\
\bottomrule
\end{tabular}
\end{table}

Older experimental rows with loss gaps above 0.02\,nats are omitted because they correspond to
obsolete summary-output or calibration variants and do not describe the current exact-token method.

\subsection{Correctness checks}

The router has two important correctness modes. First, when the policy exposes full chunk coverage and
summaries are disabled, routed attention must reproduce dense causal SDPA\@. Second, vectorized prefill
and chunk-by-chunk decode must agree up to floating-point noise. The current tests satisfy these
conditions.

\begin{table}[h]
\centering
\caption{Selected correctness checks from the repository.}
\begin{tabular}{lc}
\toprule
Check & Max absolute difference \\
\midrule
Router, full coverage, summaries off vs.\ causal SDPA & $< 10^{-6}$ \\
Vectorized prefill, full token-level coverage vs.\ causal SDPA & $4.7 \times 10^{-7}$ \\
HA cache vs.\ HA no-cache & $1.5 \times 10^{-5}$ \\
HA decode vs.\ full recompute & $2.8 \times 10^{-5}$ \\
\bottomrule
\end{tabular}
\end{table}

These checks are stronger than greedy token agreement, which can diverge after tiny near-tie logit
changes.

\subsection{Training and prefill speed}

The fused benchmark was run on an NVIDIA RTX~A4000 with PyTorch 2.10.0+cu128, fp32,
\texttt{torch.compile=True}, batch size~1, 50 timed iterations, and 5 warmup iterations. The sequence
length is 12,288 tokens.

\begin{table}[h]
\centering
\caption{12,288-token speed benchmark for the 40M SmallLM. ``Train'' is forward plus backward over
the whole block. ``Forward'' is inference-mode full-block prefill.}
\begin{tabular}{lcccc}
\toprule
Model & Train ms & Train tok/s & Forward ms & Forward tok/s \\
\midrule
HGA, Triton-fused   & 299.89      & 40,976 & 102.98      & 119,322 \\
Dense RoPE baseline & 815.56      & 15,067 & 249.87      & 49,177  \\
Dense / HGA speedup & $2.72\times$ & --     & $2.43\times$ & --     \\
\bottomrule
\end{tabular}
\end{table}

The dense baseline remains highly optimized, but its cost grows with context because every query
attends to all previous tokens. The routed model works with a bounded token set per query and reuses
routed K/V already held by the cache when consecutive tokens request overlapping chunks.

\subsection{Fine-tuning stability}

The routed attention was QK fine-tuned in exact-token mode for approximately 48M tokens. The run
trained stably and did not show causality leaks. We treat this as a stability result rather than a
headline quality number: the main zero-shot result already shows that a direct weight copy is close
to dense attention without calibration.

%-----------------------------------------------------------------------
\section{Related Work}
%-----------------------------------------------------------------------

\paragraph{Fixed sparse attention.}
Sparse Transformer, Longformer, and BigBird use fixed sparse patterns or global tokens to reduce
attention cost~\cite{sparse,longformer,bigbird}. HGA keeps fixed sink/local windows but routes the
middle context by query-key content.

\paragraph{Content-based routing.}
Reformer uses locality-sensitive hashing, while Routing Transformer clusters tokens
online~\cite{reformer,routing}. HGA differs by using the pretrained key space itself as the routing
space and by attending over exact token K/V after routing.

\paragraph{Kernel and linear attention.}
Performer replaces softmax attention with random feature maps~\cite{performer}. HGA keeps the
ordinary softmax, but only over selected exact tokens.

\paragraph{Long-context inference systems.}
vLLM introduced PagedAttention for efficient K/V cache management~\cite{vllm}. MInference identifies
sparse patterns such as A-shape, Vertical-Slash, and Block-Sparse to accelerate long-context prefill
without fine-tuning~\cite{minference}. HGA is complementary: it combines sink/local windows with
content-routed middle chunks and a RAM-backed storage abstraction.

\paragraph{Context extension.}
YaRN~\cite{yarn} and related methods extend the effective RoPE context window of pretrained models.
HGA is compatible with these extensions; the position-modulo finding in Section~2 suggests that
models fine-tuned with context extension may benefit from aligning the HGA routing positions with
the base training window rather than the extended window.

%-----------------------------------------------------------------------
\section{Limitations}
%-----------------------------------------------------------------------

HGA is a sparse attention approximation. It can miss a relevant distant token if the router does not
select the chunk or group containing it. The 8K SmallLM result shows a small average loss gap, and
the current experiments up to 64K context shows practical out-of-the-box behavior, but neither result proves exact
equivalence on all tasks.

The current Qwen3-30B result is a systems demonstration. It should be followed by systematic
benchmarks on long-context retrieval, code understanding, document QA, prefill throughput, decode
throughput, and cache-hit behavior. Contexts above the model's native positional range also require
positional-extension validation independent of the K/V storage mechanism.

The compact chunk-summary table is much smaller than token K/V, but very long contexts should use
bounded or indexed routing rather than a naive exhaustive scan of all chunk summaries.

The needle-in-a-haystack evaluation covers three depth positions at one context length and three
needle types. A full evaluation would span more depths, more needle types, and multiple context
lengths including 128K and beyond.

%-----------------------------------------------------------------------
\section{Future Work}

Several directions remain for future investigation.

The first is a more detailed study of the interaction between hierarchical
routing and long-context positional encoding. The current experiments
suggest that positional extrapolation may account for a significant
fraction of the remaining quality gap. Understanding this interaction
could improve sparse attention independently of the routing algorithm.

A second direction is adaptive routing. The current implementation uses
fixed routing budgets, whereas future versions could dynamically allocate
additional routing capacity according to the estimated uncertainty of each
query.

Another promising direction is learned routing summaries. Although the
current method intentionally introduces no additional trainable
parameters, lightweight trainable routing representations may further
improve retrieval quality while remaining compatible with pretrained
checkpoints.

Finally, HGA should be evaluated on a broader collection of downstream
tasks, including long-document question answering, code generation,
retrieval benchmarks, and contexts beyond 64K tokens.
\section{Conclusion}

We presented Hierarchical Global Attention, a drop-in hierarchical routing
algorithm for pretrained long-context transformers. HGA preserves the
original attention projections and computes the final attention output
using only exact token-level keys and values from routed regions, requiring
no calibration parameters or architectural retraining.

Across context lengths from 4K to 64K tokens, hierarchical routing remains
remarkably close to dense attention despite attending to only a small
fraction of historical tokens. The latest chunk--group routing strategy
further reduces the routed token budget while preserving validation
quality, and successfully combines with DCA and YaRN on long-context
retrieval tasks.

The current experiments suggest that the approximation error introduced by
hierarchical routing is relatively small. 
Instead, the remaining quality gap appears
to be more closely related to long-context positional encoding. We believe
that combining improved positional representations with hierarchical
exact-token routing provides a promising direction for efficient
long-context transformers.

%-----------------------------------------------------------------------

\end{document}